\documentclass[letterpaper, 10 pt, journal, twoside]{IEEEtran}
\usepackage{amsmath,amssymb,amsfonts}
\usepackage{graphicx}
\usepackage{bm}
\usepackage{subfig}
\usepackage[font=footnotesize]{caption}
\usepackage[ruled,linesnumbered]{algorithm2e}
\SetKwInput{KwInput}{Input}                
\SetKwInput{KwOutput}{Output}              
\usepackage{cite}
\usepackage{balance}
\usepackage{comment}
\usepackage{hyperref}
\usepackage{soul}
\usepackage{xcolor}
\soulregister\cite7
\soulregister\ref7
\usepackage{makecell}
\newcommand{\R}{\mathbb R}

\begin{document}

\title{\LARGE \bf Event-Triggered Model Predictive Control with Deep Reinforcement Learning for Autonomous Driving}

\author{Fengying Dang$^{1}$, Dong Chen$^{2}$, Jun Chen$^{3}$ and Zhaojian Li$^{1*}$%

\thanks{This work is supported by National Science Foundation grant \#2045436.}
\thanks{$^1$ Fengying Dang and Zhaojian Li are with the Department of Mechanical Engineering at Michigan State University, East Lansing, MI 48824, USA.
{Email: \textit{dangfen1@msu.edu, lizhaoj1@egr.msu.edu}}}%
\thanks{$^2$ Dong Chen is with the Department of Electrical and Computer Engineering at Michigan State University, East Lansing, MI 48824, USA. {Email: \textit{chendon9@msu.edu}}}%
\thanks{$^3$ Jun Chen is with the Department of Electrical and Computer Engineering at Oakland University, Rochester, MI 48309, USA. {Email: \textit{junchen@oakland.edu}}}%
\thanks{$*$Zhaojian Li is the corresponding author.}
}

\maketitle

\begin{abstract}
Event-triggered model predictive control (eMPC) is a popular optimal control method with an aim to alleviate the computation and/or communication burden of MPC. However, it generally requires \textit{priori} knowledge of the closed-loop system behavior along with the communication characteristics for designing the event-trigger policy. This paper attempts to solve this challenge by proposing an efficient eMPC framework and demonstrate successful implementation of this framework on the autonomous vehicle path following. First of all, a model-free reinforcement learning (RL) agent is used to learn the optimal event-trigger policy without the need for a complete dynamical system and communication knowledge in this framework. Furthermore, techniques including prioritized experience replay (PER) buffer and long-short term memory (LSTM) are employed to foster exploration and improve training efficiency. In this paper, we use the proposed framework with three deep RL algorithms, i.e., Double Q-learning (DDQN), Proximal Policy Optimization (PPO), and Soft Actor-Critic (SAC), to solve this problem. Experimental results show that all three deep RL-based eMPC (deep-RL-eMPC) can achieve better evaluation performance than the conventional threshold-based and previous linear Q-based approach in the autonomous path following. In particular, PPO-eMPC with LSTM and DDQN-eMPC with PER and LSTM obtains a superior balance between the closed-loop control performance and event-trigger frequency. The associated code is open-sourced and available at: \url{https://github.com/DangFengying/RL-based-event-triggered-MPC}.
\end{abstract}

\section{INTRODUCTION}
\IEEEPARstart{A}{utonomous} vehicles have attracted researchers' attention dramatically in recent years due to the advanced technology in automation, high-speed communication network and new energy. Path planning and path following are two major tasks for the behaviour control of autonomous vehicles~\cite{ayawli2018overview,muraleedharan2021real}. Path planning is executed to plan the path considering safety constraints, and a controller is then used to follow this path accurately by considering the current states and providing  suitable control. Path planning has been well explored by many researchers~\cite{liu2017path,dolgov2010path}. However, path following still remains a problem due to the high dynamic, limited computation and communication of autonomous vehicles. The path following controller are expected to provide accurate control inputs in real-time with constrained computation and communication. Path following control can be implemented using different controllers, e.g., proportional–integral–derivative (PID) control, state feedback controllers, model predictive control (MPC), and so on.

MPC is capable of handling multi-input multi-output (MIMO) systems with various  constraints, making it specially suitable for real-world autonomous vehicle path following problem. MPC can be dated back to the 1980s when engineers in the process industry first began to deploy it in real-world practice~\cite{garriga2010model}. Since then, the increasing computing power of microprocessors has greatly improved its application scope~\cite{o2016introduction}. MPC uses a system model to predict its future behavior, and selects the best control action by solving an optimization problem~\cite{baumann2018deep,baurnann2019event,hosseinloo2020event,sedghi2020machine,yoo2021event}. Despite the advances of MPC over the years~\cite{mammarella2020computationally,liu2020computationally,ding2018model,li2021review,serale2018model}, solving the constrained optimal control problem requires high computational power, which is further increased as the system dimension and prediction horizon increase. This has hindered its application to autonomous vehicles' path following that require a short sampling time but have limited computation power. 
To reduce computational burden without significantly degrading control performance, event-triggered MPC (eMPC) has emerged as a promising paradigm where MPC algorithm is solved -- instead of at each time instant as in the traditional MPC implementation -- only when triggered by a predefined trigger condition   \cite{brunner2017robust,li2014event,chen2021comparison,he2015event,eqtami2011novel,huang2022event}. In such framework, a triggering event can be defined based on either the deviation of the system states~\cite{brunner2017robust,li2014event,chen2021comparison} or the cost function value~\cite{he2015event,eqtami2011novel}. By solving the optimization problem only when necessary, eMPC can significantly reduce  online computations. However, the trigger mechanism design, concerning when to trigger the optimization so as to preserve system performance while keeping the number of triggers low, still remains a challenge~\cite{sedghi2020machine}.


The most common event-trigger policy is the threshold-based event-trigger policy, where an event is triggered if the predicted state trajectory and real-time feedback diverge beyond a certain threshold \cite{brunner2017robust,li2014event,chen2021comparison}. However, the threshold calibration is usually based on the knowledge of the closed-loop system behavior which is not always available, especially for complex systems. To address this limitation, our prior work \cite{chen2022reinforcement} investigates the use of model free RL techniques, a simple linear Q-learning approach, to synthesize a triggering policy with the aim of achieving the optimal balance between control performance and computational efficiency. However, this linear Q-learning has a hard time capturing the nonlinear event-trigger policy, leading to unnecessarily high event frequency. Therefore, in this paper, we propose to use deep RL to learn the event-trigger policy which makes the proposed framework achieve better trade-offs between system performance and computation cost.

This paper addresses the autonomous driving path following problem using a novel eMPC framework. First, it extends the previous work \cite{chen2022reinforcement} with an improved vehicle model, thereby removing the limitation of using only the front steering angle as driving control. Second, we develop a \textit{model-free} deep-RL-eMPC framework that uses deep RL to learn the event-trigger policy online, so that no prior knowledge of the closed-loop system is needed, which is essential for a dynamic and complex system. Both off-policy and on-policy RL methods are tested. Meanwhile, techniques including prioritized experience replay (PER) buffer and long-short term memory (LSTM) are exploited to significantly improve the training efficiency and control performance. The validity of the proposed deep-RL-eMPC is demonstrated using a nonlinear autonomous vehicle model and the results show that our approach clearly outperforms the conventional threshold-based approach in \cite{chen2021comparison} and the previous linear Q-learning based approach in \cite{chen2022reinforcement}.

The remainder of the paper is organized as follows. Section \ref{sec:review-RL-MPC}
reviews relevant literature on RL and MPC integration, to provide more context for the presented work. Section~\ref{sec1+:PROBLEM FORMULATION} formulates the autonomous vehicle path following problem. Section~\ref{sec:RLeMPC framework} presents the framework of eMPC with triggering policy obtained from RL. The experiment setup and results of the proposed deep-RL-eMPC method in the autonomous vehicle path following problem are presented in~\ref{sec:RLeMPC for autonomous vehicle path following}. Finally, conclusion remarks are provided in Section ~\ref{sec6:CONCLUSION}.
\section{RELEVANT WORK ON RL/MPC INTEGRATION}\label{sec:review-RL-MPC}
Utilizing RL to aid MPC is not new in literature. For example, \cite{farshidian2019deep,karnchanachari2020practical} propose an off-policy actor-critic algorithm called DMPC, where an off-policy critic learns a value function while the actor utilizes MPC to interact with the environment. It is assumed that the system dynamics are known, but the cost function that MPC should minimize is unknown and is learned by the critic's value function estimation. Both analytical and numerical results demonstrate improvements on learning convergence.

RL can also be used to learn the system dynamics that are then used by MPC for prediction \cite{cui2019reinforcement,cui2021autonomous,kuo2020sample,kamthe2018data,ostafew2016learning,shin2019autonomous,hewing2019cautious,kabzan2019learning,carron2019data}. This approach is called model-based RL in \cite{cui2019reinforcement,cui2021autonomous,kuo2020sample,kamthe2018data,ostafew2016learning,shin2019autonomous,hewing2019cautious,kabzan2019learning,carron2019data}. Specifically,
\cite{cui2019reinforcement,cui2021autonomous,kuo2020sample,kamthe2018data,ostafew2016learning} studies learning based probabilistic MPC in the framework of RL, where the system dynamics and environment uncertainties are modeled as Gaussian Process (GP), whose parameters are iteratively identified through trial and error. Authors of \cite{hewing2019cautious,kabzan2019learning,carron2019data} use a GP model to learn errors between measurement and a nominal model, which are then used to set up the optimal control problem for MPC to guarantee constraints robustness. 

Reference \cite{shin2019autonomous} combines RL and MPC in the context of surgical robot control. The system dynamics are modeled by artificial neural network (ANN), whose parameters are identified through RL or learning from demonstration. In RL approach, the agent explored the action space using $\epsilon$-greedy, collected observations, and iteratively trained the ANN to model system dynamic, while an MPC is used to optimize action based on the trained ANN. In the learning from demonstration approach, the ANN parameters are initialized using observations collected from human operators.

Finally, RL can also be used to directly optimize MPC control law. For example, \cite{soloperto2018learning} proposes a robust MPC where the control law is restricted to an affine function of the feedback with the gain being pre-computed offline and the offset being learnt by RL. 
Reference \cite{soloperto2018learning} additionally shows that the robust MPC can also reject disturbance when the Gaussian process model is unknown and learnt online. Authors in \cite{bohn2021optimization} investigated the use of gradient based Partially Observable Markov Decision Processes (POMDP) algorithm to train the RL recomputation policy for event-triggered MPC control to save energy~\cite{baxter2001infinite}. However, the solutions of POMDP algorithm often suffer from the high variance of the gradient estimate~\cite{xu2015acis}.

To the best of our knowledge, the use of deep RL to trigger MPC has not been reported in literature. In this paper, we attempt to fill this gap by investigating deep RL-based event-triggered MPC, or deep-RL-eMPC, which learns the optimal event-trigger policy without requiring any knowledge on the closed-loop dynamics and therefore significantly reduces the amount of calibrations.

\section{PROBLEM FORMULATION}
\label{sec1+:PROBLEM FORMULATION}
This paper aims to improve autonomous vehicles path following control by proposing a systematic, algorithmic framework where eMPC can be used without having the prior knowledge of the closed-loop system behavior. Our goal is to use an RL agent to learn the optimal event-trigger policy automatically.
\subsection{Task Description: Autonomous Vehicle Dynamics and Path Following Problem}
In order to demonstrate the proposed deep RL-eMPC and its improving techniques, a path following task is chosen. For a single track vehicle model, the equations for vehicle center of gravity (CG) and wheel dynamics are given by
\begin{subequations}\label{bike-nl}\allowdisplaybreaks
\begin{align}
\dot l_x &= v_x\cos\psi-v_y\sin\psi,\\
\dot v_x &= v_yr+\frac{2}{m}\sum_{i=f,r}F_{x,i}-g\sin\sigma_g-\frac{1}{m}F_a\label{bike-vx},\\
\dot l_y &= v_x\sin\psi+v_y\cos\psi,\\
\dot v_y &= -v_xr+\frac{2}{m}\sum_{i=f,r}F_{y,i}\label{bike-vy},\\
\dot \psi &= r,\\
\dot r &= \frac{1}{I}\left(2L_{x,f}F_{y,f}-2L_{x,r}F_{y,r}\right), \label{bike-r}
\end{align}
\end{subequations}
where $l_x$ and $l_y$ are the longitudinal and lateral position of the center of gravity of vehicle, respectively;  $\psi$ is the vehicle  rotational angle along the longitudinal axis in the \emph{global} inertial frame; and $v_x$, $v_y$, and $r$ are, respectively, the vehicle longitudinal velocity, lateral velocity, and yaw rate in the \emph{vehicle} frame. $F_a$ is the aerodynamic drag force~\cite{rajamani2011vehicle} and $F_x$ and $F_y$ are tire forces. $m$ is the vehicle mass, $I$ is the vehicle rotational inertia on yaw dimension, $L_{xf}$ and $L_{xr}$ are the distance from CG to the middle of front and rear axle, respectively.

The tire force $F_{x,i}$ and $F_{y,i}$ in (\ref{bike-vx}), (\ref{bike-vy}) in vehicle frame can be modeled by
\begin{subequations}\label{equ-fx-fy}
\begin{align}
F_{x,i} &= \bar F_{x,i}\cos\beta_i-\bar F_{y,i}\sin\beta_i\\
F_{y,i} &= \bar F_{x,i}\sin\beta_i+\bar F_{y,i}\cos\beta_i,
\end{align}
\end{subequations}
where $\beta_i$ is the wheel-road-angle for the wheel $i$, $i=\{f,r\}$ represents the front or rear wheel, $\bar F_{x,i}$ and $\bar F_{y,i}$ are the tire force in wheel frame which can be obtained as
\begin{subequations}
\begin{align}\label{equ-fxfy-bar}
\bar F_{x,i} &= \frac{T_i}{2R}, \quad\quad\quad \\
\bar F_{y,i} &= C_i \mu_i F_{z,i}\alpha_i,
\end{align}
\end{subequations}
where $T_i$ is the propulsion/braking torque along the axle, $R$ is the effective tire radius, $C_i$ is the tire corner stiffness and $\mu_i$ characterize the road surface, $\alpha_i$ is the slip angle. We refer readers to \cite{chen2021comparison} for a detailed computation of the slip angle $\alpha_i$.

The normal force $F_{z,i}$ in (\ref{bike-r}) can be modeled by static load transfer,
\begin{equation}\label{equ-fz}
F_{z,i} = \frac{L_{x,i}mg}{2(L_{x,f}+L_{x,r})}.
\end{equation}

 In this paper, we consider a problem of autonomous vehicle following a sinusoidal trajectory using the proposed deep-RL-eMPC method~\cite{kong2015kinematic,chen2021comparison}, the following path is given by
\begin{equation}\label{equ-ref}
l_y = g(l_x)=4\sin\left(\frac{2\pi}{100} l_x \right).
\end{equation}
\subsection{Optimal Control Problem and its Goal}
Consider a discrete-time system with the following dynamics 
\begin{equation}\label{mpc-nmodel}
x_{t+1} = f(x_t,u_t),
\end{equation}
where $x_t\in \R^n$ is the system state at discrete time $t$ and $u_t\in\R^m$ is the control input. Given a prediction horizon $p$, MPC aims to find the optimal control sequence $U_t$ and optimal state sequence $X_t$ by solving the following optimal control problem:
\begin{subequations}\label{equ-nmpc-ocp}
\begin{align}
\min_{X_t,U_t} \quad & J_{\text{mpc}}=\sum_{k=0}^p \ell(x_{t+k},u_{t+k})\label{equ-cons-J}\\
\textrm{s.t.} \quad & x_{t}=\hat x_{t}\label{equ-cons-zeta0}\\
  & x_{t+k}=f(x_{t+k-1},u_{t+k-1}), \quad 1\leq k\leq p\\
  & x_{min}\leq{x}_{t+k}\leq{x}_{max}, \quad 1\leq k\leq p\label{equ-cons-x}\\
  & u_{min}\leq u_{t+k}\leq u_{max}, \quad 0\leq k\leq p-1\label{equ-cons-u}\\
  & \Delta_{min}\leq u_{t+k}-u_{t+k-1}\leq \Delta_{max}, \nonumber\\
  &\quad\quad\quad\quad\quad\quad\quad\quad\quad\quad 0\leq k\leq p-1, \label{equ-cons-du}
\end{align}
\end{subequations}
where $U_t$ and $X_t$ are defined as $U_t = \{u_t,u_{t+1},\dots,u_{t+p-1}\}$ and $X_t = \{x_{t+1},x_{t+2},\dots,x_{t+p}\}$, $\ell(x_{t+k},u_{t+k})$ is the stage cost function, $\hat{x_t}$ denotes the real state or current state estimation, and $u_{t+k}$ denotes the control action at time step $t+k$. For conventional time-triggered MPC, the above optimal control problem is solved for every sampling time $t$, and only the first element $u_t$ of $U_t$ is applied to the system as the control command, while all the remaining elements $u_{t+1},\dots,u_{t+p-1}$ are abandoned.


 Let $t$ and $t_p$ represent the current time step and the last event time, respectively, and there thus exists a $k\in\mathbb{N}$ such that  $t=t_p+kdt$ where $dt$ is the sampling time of the discrete system. Let $a_t$ denotes the triggering command in event-triggered MPC at time step $t$. Then when $a_t=1$, the above optimal control problem is solved and the first element of the optimal control sequence $U_{t}$ computed at current time step $t$ will be used as control command. When $a_t=0$, the optimal control sequence $U_{t_p}$ computed at last event when the time instance equals to $t_p$ will be shifted to determine the control command~\cite{chen2021comparison}. Then the control input $u$ can be compactly represented as:
\begin{equation}\label{equ-u-non-e}
u_t = \left\{\begin{array}{ll} 
U_{t}(1), & \quad
\text{if} \quad a_t=1,\\ 
U_{t_p}(k+1), & \quad 
\text{if} \quad a_t=0, 
\end{array}\right.
\end{equation}

\begin{figure}[tpb]
	\centering
	\includegraphics[width=1\linewidth]{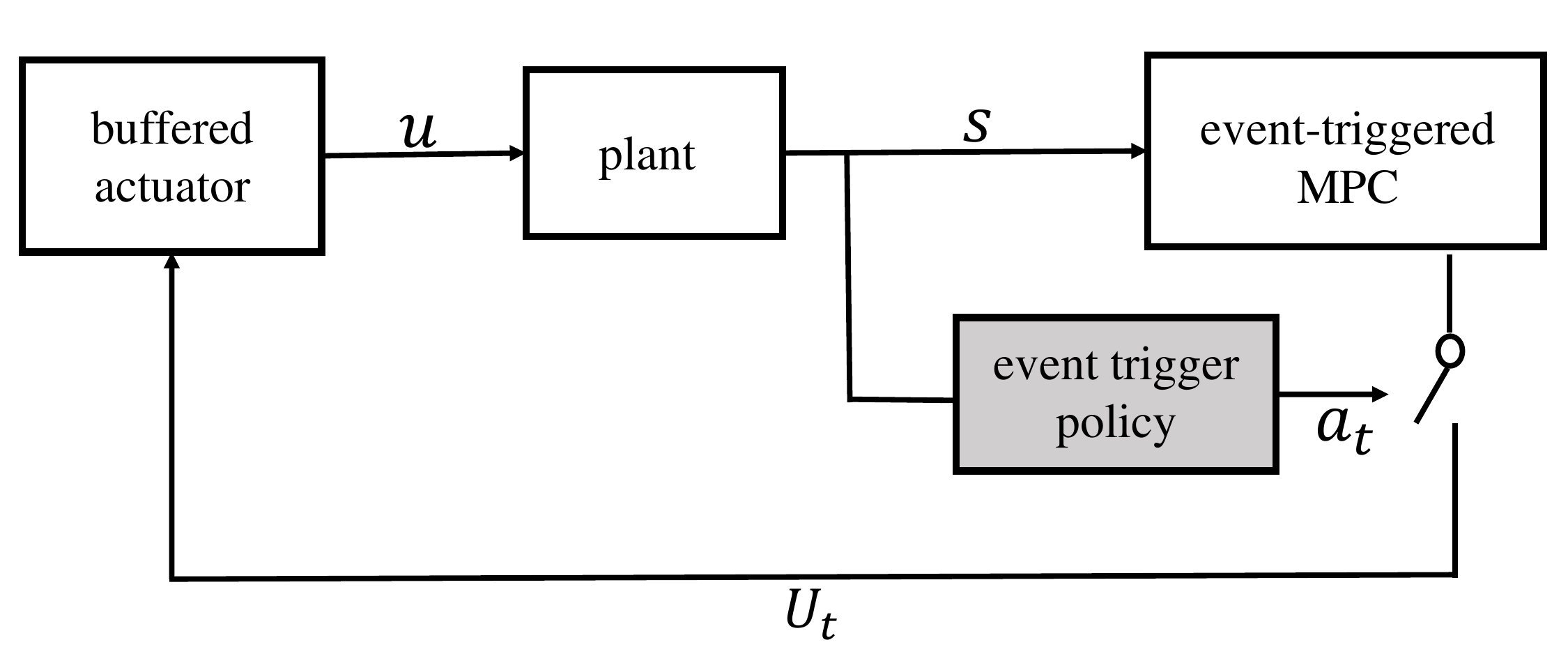}
	\caption{The scheme of event-triggered model  predictive  control (eMPC).}
	\label{fig: eMPC}
\end{figure}

To implement (\ref{equ-u-non-e}) for eMPC, a buffer can be used to store the optimal control sequence $U_{t_p}$ computed at last event at time $t_p$. At each time step, the event-trigger policy block generates $a_t$ based on current feedback from the plant. In eMPC, only when $a_t=1$, a new control sequence $U_t$ is computed by solving (\ref{equ-nmpc-ocp}), whose first element is implemented by actuator as $u$, while the entire sequence is saved into buffer. If $a_t=0$, indicating the absence of an event, the control sequence currently stored in the buffer will be shifted based on the time elapsed since last event to determine the current control input $u$. This process is depicted in Fig.~\ref{fig: eMPC}. 

In general, the event $a_t$ can be generated by certain event-trigger policy $\pi$, denoted as,
\begin{equation}\label{equ-e}
a_t \sim \pi_{\theta}(X_{t_p},\hat{x}_t),
\end{equation}
where $X_{t_p}$ is the optimal state sequence computed at last event when $a_{t_p}=1$ and $\hat{x}_t$ is the real state (or current state estimate if not directly measured), $\theta$ are parameters characterizing the policy. It is worth noting that, for nonlinear constrained MPC, the design of event-trigger policy $\pi$ is challenging and requires extensive calibration and prior knowledge of the closed-loop system behavior. Therefore, the design of event-trigger policy and its calibrations are usually problem specific and non-trivial. To address this limitation, the objective of this paper is to learn the optimal event-trigger policy $\pi$ using model-free deep RL techniques. 

If we discretize (\ref{bike-nl}) to obtain a discrete-time model in the form of (\ref{mpc-nmodel}), with $x = \left[l_x,v_x,l_y,v_y,\psi,r\right]$ and $u = \left[T_f,\beta_f \right]$ where $T_f$ is the axle driving torque and $\beta_f$ is the front steering angle. The stage cost of (\ref{equ-cons-J}) is defined as
\begin{equation}\label{equ-mpc-stage}
\ell(x,u)=\left|\left|x(3)-4\sin\left(\frac{2\pi}{100}x(1)\right)\right|\right|^2_{Q_t}+||u-u^r||^2_{Q_u},    
\end{equation}
where the first nonlinear term penalizes the path tracking error and the second term penalizes large control efforts. Here the norm is defined as $\left|\left|x\right|\right|^2_Q=x^T Q x$. More specifically, the MPC cost function $J_{\text{mpc}}$ in~(\ref{equ-cons-J}) in this case can be equivalently represented as:
\begin{align}\label{equ-cons-J-nmpc}
J_{\text{mpc}}(X_t,U_t)&=\sum_{k=1}^p\left|\left|x_{t+k}(3)-4\sin\left(\frac{2\pi}{100}x_{t+k}(1)\right)\right|\right|^2_{Q_t}\nonumber\\
& \quad\quad\quad\quad\quad\quad\quad +\sum_{k=0}^{p-1}\left(||u_{t+k}-u_{t+k}^r||^2_{Q_u}\right),
\end{align}
where $U_t$ and $X_t$ are defined as $U_t = \{u_t,u_{t+1},\dots,u_{t+p-1}\}$ and $X_t = \{x_{t+1},x_{t+2},\dots,x_{t+p}\}$, and the terms independent of $X_t$ and $U_t$ are ignored. 

\section{ EVENT-TRIGGERED MPC WITH DEEP RL-BASED POLICY LEARNING}
\label{sec:RLeMPC framework}

In this section, we present our proposed deep RL-based policy learning eMPC, or deep-RL-eMPC.
\subsection{Deep-RL-eMPC Framework}
\begin{figure}[tpb]
	\centering
	\includegraphics[width=1\linewidth]{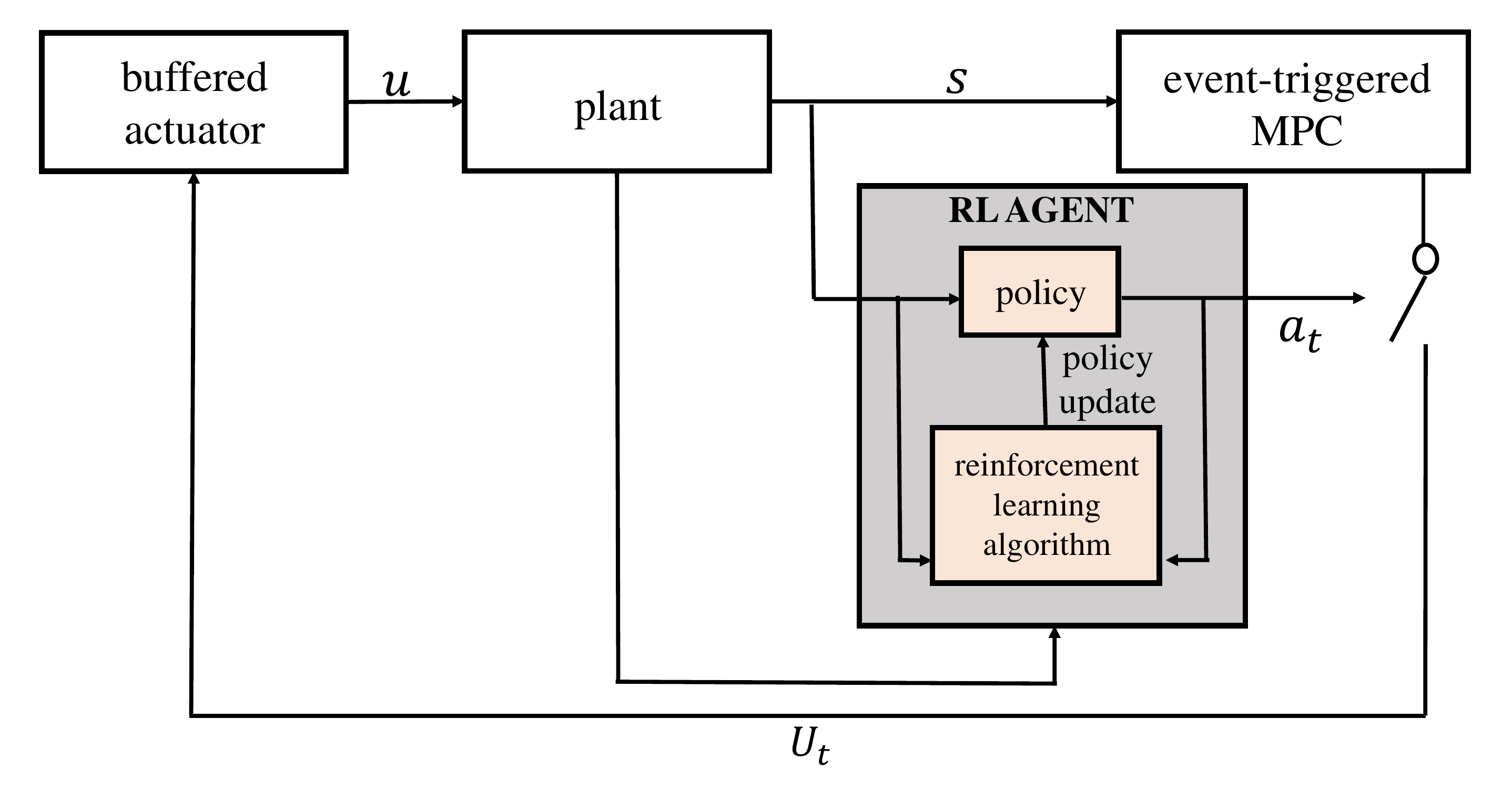}
	\caption{The scheme of RL based event-triggered MPC.}
	\label{fig: RLeMPC}
\end{figure}

The process of our deep-RL-eMPC framework is shown in Fig.~\ref{fig: RLeMPC}. The RL agent learns the event-trigger policy parameter $\theta$ by continuously interacting with the environment. Specifically, at each time step, the agent sends an action $a$ to the environment. The environment then implements the eMPC following~(\ref{equ-u-non-e}), simulates the dynamic system following~(\ref{mpc-nmodel}), and emits an immediate reward following the designed reward function. The agent then observes the reward signals, update $\theta$, and transitions to next state.

For an eMPC problem, the discrete action space for RL agent is defined as $\mathcal{A}=\{0,1\}$, where the event will be triggered when $a=1$ and will not be triggered when $a=0$. 
As the feedback from the environment, the immediate reward function is defined as
\begin{equation}\label{r-nmpc}
r_t \triangleq -\ell(\hat x_t,u_t)dt-\rho_c a_t,
\end{equation}
where the first term $\ell(\hat x_t,u_t)dt$ measures the closed-loop system performance and the second term $\rho_c a_t$ measures the cost of triggering events. Note that $\ell(\hat x_t,u_t)$ is the stage cost and is computed using the the real state (or current state estimate if not directly measured) $\hat x_t$ and real-time control (\ref{equ-u-non-e}). Furthermore, $\rho_c$ is a hyper-parameter used to balance between control performance index and triggering frequency. One can fine tune this hyperparameter $\rho_c$ to make a tradeoff between control performance and computational cost.

The complete deep-RL-eMPC algorithm is shown in Algorithm$~$\ref{PSEUDO CODE: RL-based Event-Triggered MPC}. In this algorithm, $M$ is the total number of training epochs, $T$ is the length of each episode representing total training time in each epoch, $\gamma$ is the discount factor in the reward function, $dt$ is the discrete time step, and $N$ is the size of sampled experiences at each time (batch size). The output of Algorithm$~$\ref{PSEUDO CODE: RL-based Event-Triggered MPC} is the system parameters $\theta$. The RL agent interacts with the environment for $M$ number of epochs (Lines 2-24). After initialization, Lines 5 shows how to choose action. Lines 7-12 implement the event-triggered MPC to compute the control command $u$, which is used to simulate the dynamical system (\ref{mpc-nmodel})  (Line 13). After that, the environment emits next state $s_{t+1}$ and immediate reward $r_t$ (Lines 16), which is observed by RL agent (Line 18). The latest experience tuple $(s_t,a_t,r_t,s_{t+1})$ is then added into an experience buffer ${\cal D}$  (Line 19). The RL parameters $\theta$ is updated using a batch of $N$ experiences sampled from the experience buffer ${\cal D}$ (Line 20). RL agent then moves to next state (Lines 21). After each epoch, RL agent is reset for the next epoch (Line 3). Lines 7-16 are part of the environment, whose computation is unknown to the RL agents. Note that the agent only observes the environment outputs, i.e., next state and reward.

\begin{algorithm}[ptb]
  \caption{RL-based Event-Triggered MPC}
  \label{PSEUDO CODE: RL-based Event-Triggered MPC}
  \KwInput{$M$, $T$, $dt$, $\gamma$, $N$}
  \KwOutput{$\theta$}
  Initialize $\theta$, ${\mathcal{D}} \gets \emptyset$\;
  \For{$j=0$ to $M-1$}{
   Initialize $s_t$, $Z$, $U$, $k\gets0$\;
  \While{$t<=T$}{
    {select action $a_t \sim \pi_{\theta}(X_{t_p},\hat{x_t})$} \\
    \emph{\% Simulate Environment}\;
    \eIf{$a_t=1$}{
     $k\gets 0$\;
     $(Z,U)\gets $ Solving optimal control problem (\ref{equ-nmpc-ocp})\;}
    {$k\gets k+1$\;}
    $u\gets U(k)$\;
    $x_{t+1}\gets$ Simulate system dynamics (\ref{mpc-nmodel}) using $u$\;
    $s_{t+1}\gets(x_{t+1},Z(k))$\;
    $r_t\gets(\ref{r-nmpc})$\;
    \emph{\% End of Environment Simulation}\;
    Observe $r_t$ and $s_{t+1}$\;
    Update ${\cal D}$ to include $(s_t,a_t,r_t,s_{t+1})$\;
    Sample $N$ experiences from ${\cal D}$ and update $\theta$\; 
    $s_t\gets s_{t+1}$\;
    $t\gets t+dt$
    }
    }
Note: ${\cal D}$ can be either conventional on-policy or off-policy experience buffer or priority experience buffer.
\end{algorithm}

\subsection{Deep RL Algorithms and Improving Technique}

The framework shown in Fig. \ref{fig: RLeMPC} and Algorithm \ref{PSEUDO CODE: RL-based Event-Triggered MPC} is a general frame which can accommodate different RL algorithm. In this paper, we investigate three different RL agents, including  Double Q-learning (DDQN) \cite{mnih2015human} and Proximal Policy Optimization (PPO)\cite{schulman2017proximal}, Soft Actor-Critic (SAC) \cite{haarnoja2018soft}, and show the proposed framework is also suitable for other RL algorithms. 

In this subsection, we first briefly describe these three deep RL algorithms. Then two improving technique for Rl agent including Prioritized Experience Replay (PER) and Long Short-term Memory (LSTM) are presented.
\subsubsection{Double Q-learning}
Deep Q network is a type of Q-learning which uses neural network as a policy. To address the issues of overestimation of Q values in deep Q network \cite{mnih2013playing}, Double Q-learning (DDQN) explicitly separates action selection from action evaluation which allows each step to use a different function approximator and shows a better overall approximation of the action-value function~\cite{mnih2015human}. DDQN improves deep Q network by replacing the target $y^{DQN}$ by $y^{DDQN} = r_t+ \gamma Q_{\theta'}(s_{t+1}, \arg\max_{a} Q_\theta(s_{t+1},a))$, resulting in the Double Q-learning loss:
\begin{equation}\label{eqn:DDQN}
L_{DDQN}(\theta) = E_{\mathcal{D}}[y^{DDQN} - Q_{\theta}(s_t,a_t)]^2.
\end{equation}
\subsubsection{PPO} PPO, an on-policy policy gradient RL algorithm, replaces the KL-divergence used in TRPO \cite{schulman2015trust} with a clipped surrogate objective function (\ref{eq:ppo}), which is proved to be better suited for the TRPO and easy to implement.
\begin{equation}\label{eq:ppo}
L_{PPO}^{CLIP}(\theta) = E_t[\min{(r_t A_t, clip(r_t, 1-\varepsilon, 1+\varepsilon)A_t)}].
\end{equation}

\subsubsection{Soft Actor-Critic} SAC achieves the state-of-the-art performance across a wide range of continuous-action control problems and updates the stochastic actor-critic policy in an off-policy way. SAC takes a good exploration-exploitation trade-off via entropy regularization.

In this paper, we adopt SAC and PPO to the discrete action space setting following the discrete categorical distribution design in \cite{christodoulou2019soft}. For details, refer to DDQN~\cite{mnih2015human}, PPO~\cite{schulman2017proximal} and SAC~\cite{haarnoja2018soft}.

The training performance of the proposed deep-RL-eMPC framework depends on the quality of the selected experience sample, so how to choose them is critical when using off-policy RL algorithms. The experience replay buffer utilizes a fixed-size buffer that holds the most recent transitions collected by the policy~\cite{lin1992self,fedus2020revisiting}. In RL, the weights updating and optimization of neural networks are based on the experience replay. The experience replay in the original DDQN uniformly samples the stored experience to train the network weights. However, the importance of experiences are different. Some experiences are more valuable than others in the long run and important experience should be considered more frequently. To address this problem, the prioritized experience replay has been proposed~\cite{schaul2015prioritized} to prioritize more frequent replay transitions leading to high expected learning progress, as measured by the magnitude of their TD error. Specifically, the probability of sampling transition $i$ is defined as follows:
\begin{equation}
P(i) = \frac{p_i^{\alpha}}{\sum_k p_k^{\alpha}},
\end{equation}
where ${\alpha} \in [0, 1]$ controls how much prioritization is applied; when ${\alpha} = 0$, the experience will be sampled uniformly. Here $p_i >0$ represents the priority of transition $i$, which is initialized as 1 and updated based on the TD-error $\delta_i$ during the transition.

More specifically, to alleviate the bias of the gradient magnitudes introduced by the priority replay, importance-sampling (IS) is introduced in \cite{schaul2015prioritized} as:
\begin{equation}
w_i = (\frac{1}{\mathcal{N}} \frac{1}{P(i)})^{\beta}.
\end{equation}
where $\beta$ is the hyperparameter annealing the amount of importance-sampling correction over time. $\mathcal{N}$ is size of the experience buffer. The weight $w_i$ is then used in the Q-learning updates by replacing the TD-error $\delta_i$ as $w_i \delta_i$. In practice, we can apply the PER by replacing line 24 in Algorithm~\ref{PSEUDO CODE: RL-based Event-Triggered MPC} with the designed PER scheme.

To encode the historical information in the network, a straightforward way is to feed all historical states to the RL agent, but it increases the state dimension significantly and may distract the attention of the RL agent from recent input states. To address this challenge, recurrent neural network (RNN) has been developed, which is a class of artificial neural networks that can encode and learn temporal information. Traditional RNN does not have the ability for long term memory and suffers from vanishing gradient problem. Long short-term memory (LSTM) \cite{hochreiter1997long}, a type of RNN architecture, solves this issue by using feedback connections and thus suitable for long-time series data. 
In this paper, we will explore the use of LSTM as the last hidden layer to extract representations from different state types and encode the history information. 
\section{AUTONOMOUS VEHICLE PATH FOLLOWING USING DEEP-RL-EMPC}
\label{sec:RLeMPC for autonomous vehicle path following}
In this section we apply the proposed deep-RL-eMPC to a nonlinear autonomous vehicle path tracking problem. The prediction horizon of MPC is set to $p=5$ with upper and lower bounds for all control inputs. Since autonomous vehicle requires short control sampling time but has limited onboard computation power, this nonlinear path tracking problem is a good example to demonstrate  the proposed deep-RL-eMPC. 

\subsection{RL Structure and Settings}
In this paper, we encode the input state with a one fully connected (FC) layer with 128 neurons, followed by two 128-neuron FC layers. In the LSTM design, we replace the last FC layer with a 128-unit LSTM layer. The last layer outputs two Q values corresponding to two actions, i.e., trigger and not trigger. The target network in DDQN are updated every $N_0=1000$ steps.

The state of the environment is defined to be $s=(\hat{x},\bar{x})$, where $\hat{x}$ as mentioned above is the state estimate of the dynamical system and $\bar{x}$ is the MPC prediction made at last event. The reward function follows~(\ref{r-nmpc}), with $\ell(\hat x_t,u_t)$ defined as follows:
\begin{align}
\ell(\hat x_t,u_t)=\left|\left|\hat x_t(3)-4\sin\left(\frac{2\pi}{100}\hat x_t(1)\right)\right|\right|^2_{Q_t}+||u_{t}-u_{t}^r||^2_{Q_u},
\end{align}
where $\hat x_t$ is the real state (or current state estimate if not directly measured) and $u_t$ is the real-time applied control computed by (\ref{equ-u-non-e}). Then the return for one episode in the RL algorithm is as follows:
\begin{equation}
R =\sum_{t=1}^{T_e} r_t = \sum_{t=1}^{T_e} \left(-\ell(\hat x_t,u_t)dt - \rho_c a_t \right),
\label{Eq: rhoc}
\end{equation}
where $R$ is the episodic return of RL algorithms, $T_e$ is the number of steps for the episode, $\rho_c$ is a hyper-
parameter proposed to balance control performance and event trigger frequency. To evaluate performance of different RL algorithms in our deep-RL-eMPC frame, we adopt the following two evaluation metrics: total MPC cost $E_{mpc}$ and event triggering frequency $A_f$, which are defined as follows:
\begin{align}
E_{mpc} &= \sum_{t=1}^{T_e} \left(\ell(\hat x_t,u_t)dt\right) \\
A_f &= \frac{\sum_{t=1}^{T_e} a_t}{T_e},
\end{align}

\begin{table*}[tbp]
\renewcommand{\arraystretch}{1.6}
\centering
\caption{Evaluation return R, triggering frequency $A_{\text{f}}$, and MPC cost $E_{mpc}$ using different RL agents in deep-RL-eMPC.}
\resizebox{0.9\textwidth}{!}{
\begin{tabular}{l lc cccccc} \hline
	& & Threshold & LSTDQ  & SAC & DDQN  & DDQN+LSTM+PER & PPO    & PPO+LSTM \\ \hline  \hline
	$\rho_c=0$ & return & 1.606 & 0.062 & 0.058 & 0.056 & 0.058 & \textbf{0.055} & \textbf{0.055} \\
	& $A_{\text{f}}$/$E_{mpc}$& 0.118/1.606 & 0.902/0.062 & 0.99/0.058  & 0.902/0.056 & 0.99/0.058   & 0.99/0.058  & 0.99/0.055  \\ \hline
	$\rho_\text{c}=0.001$ & return & 1.618 & 0.157 & 0.158 & 0.152 & 0.137 & 0.119 & \textbf{0.112} \\
	& $A_{\text{f}}$/$E_{mpc}$& 0.118/1.606 & 0.931/0.157 & 0.98/0.058  & 0.951/0.055 & 0.794/0.056  & 0.594/0.059 & 0.594/0.059 \\ \hline
	$\rho_\text{c}=0.01$ & return & 1.728 & 0.66  & 1.015 & 0.627 & \textbf{0.431} & 0.634 & 0.529 \\
    & $A_{\text{f}}$/$E_{mpc}$& 0.118/1.606 & 0.559/0.660 & 0.922/0.075 & 0.5/0.117 & 0.255/0.171  & 0.515/0.114 & 0.515/0.069 \\ \hline
    \end{tabular}}
	\label{table:Evaluation return}
\end{table*}

\begin{figure*}[]
    \hspace*{0cm}  
	\centering
	\includegraphics[width=0.85\linewidth]{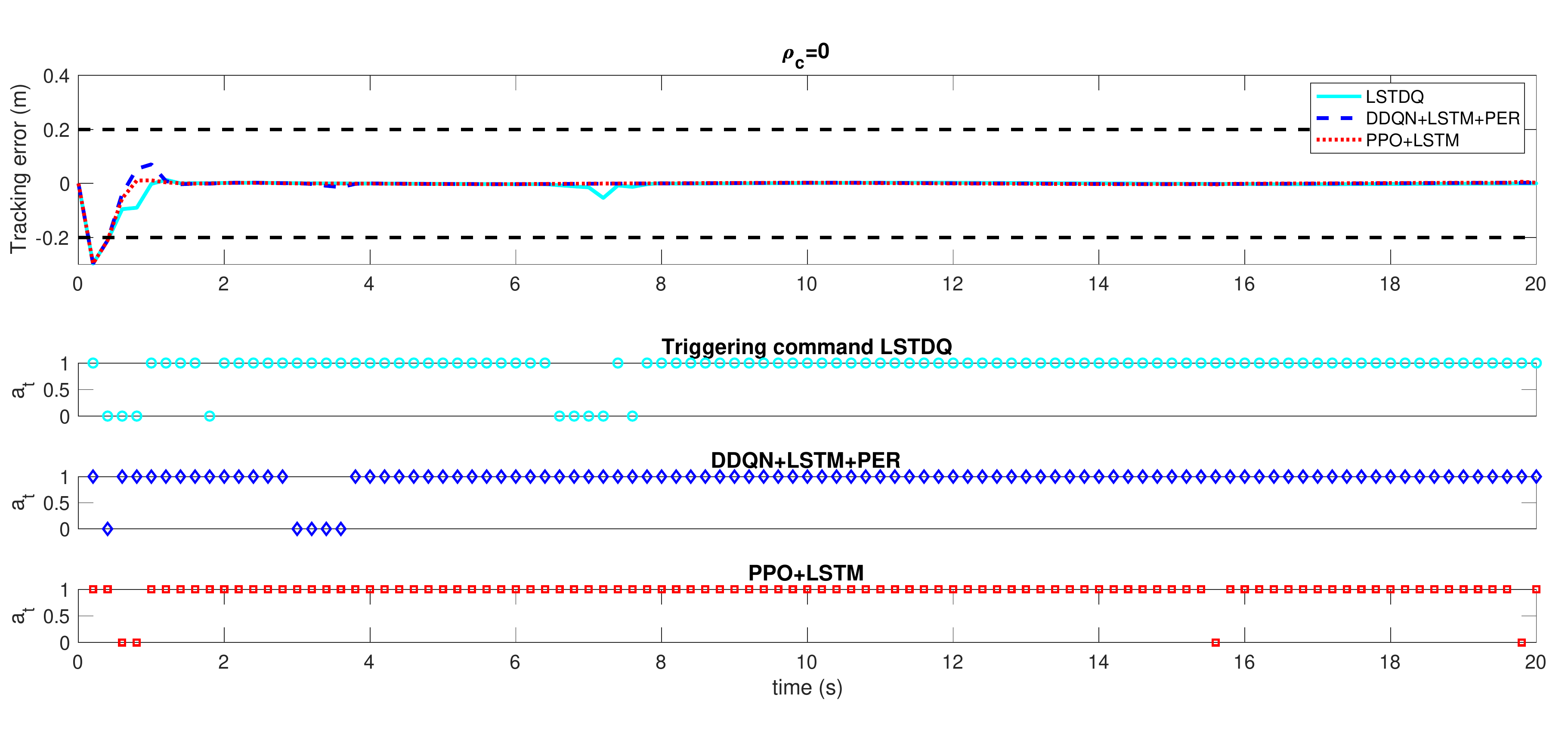}
	\caption{Experimental results of deep-RL-eMPC for the reward function with $\rho_c=0$. The comparison of the tracking error using three different RL algorithm in deep-RL-eMPC (first row). The corresponding triggering commands $a_t$ during the process when using LSTDQ (second row), DDQN+LSTM+PER (third row) and PPO+LSTM (fourth row).} 
	\label{fig: LQ}
\end{figure*}

\begin{figure*}[]
    \hspace*{0cm}  
	\centering
	\includegraphics[width=0.85\linewidth]{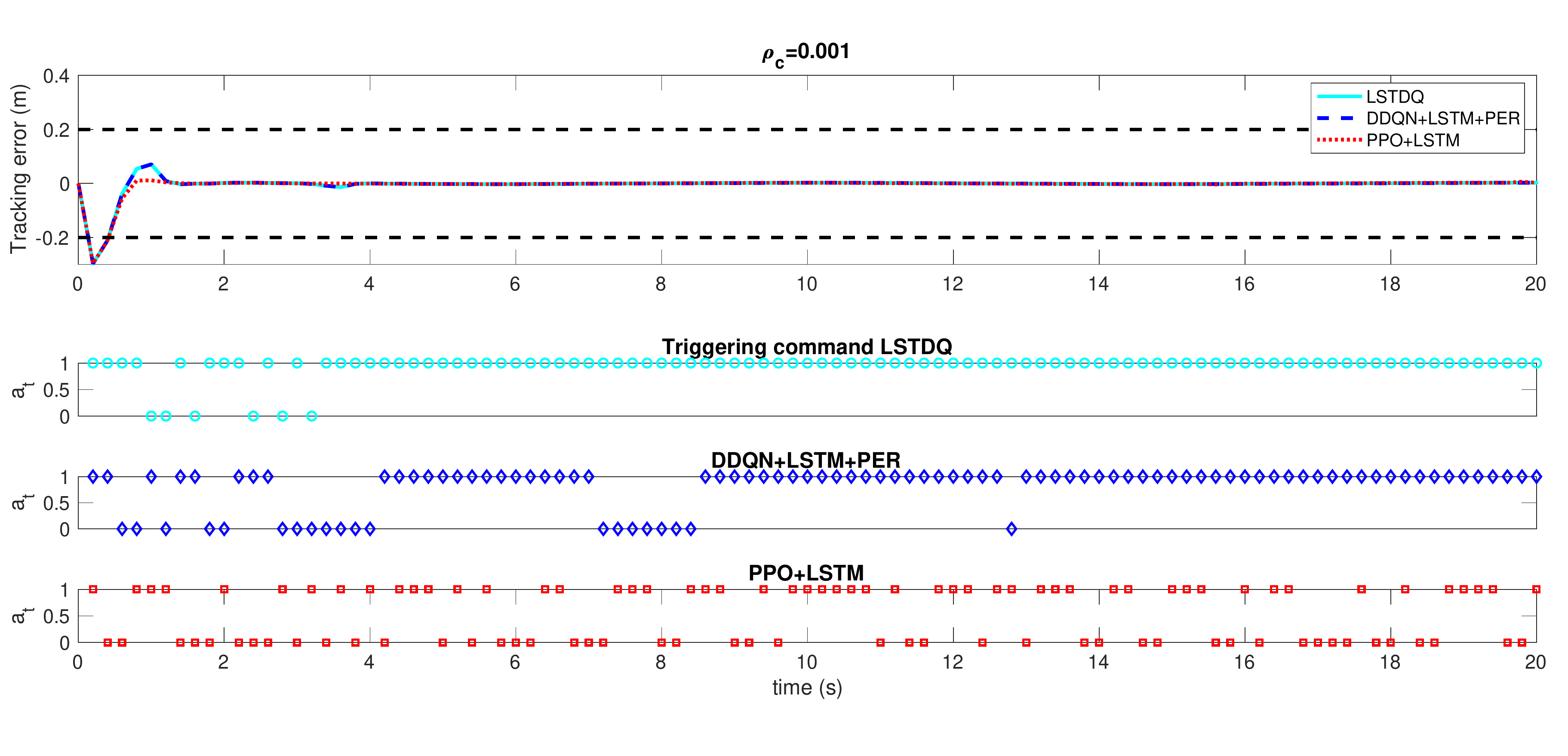}
	\caption{Experimental results of deep-RL-eMPC for the reward function with $\rho_c=0.001$. The comparison of the tracking error using three different RL algorithm in deep-RL-eMPC (first row). The corresponding triggering commands $a_t$ during the process when using LSTDQ (second row), DDQN+LSTM+PER (third row) and PPO+LSTM (fourth row).} 
	\label{fig: PPO}
\end{figure*}

\begin{figure*}[]
    \hspace*{0cm}  
	\centering
	\includegraphics[width=0.85\linewidth]{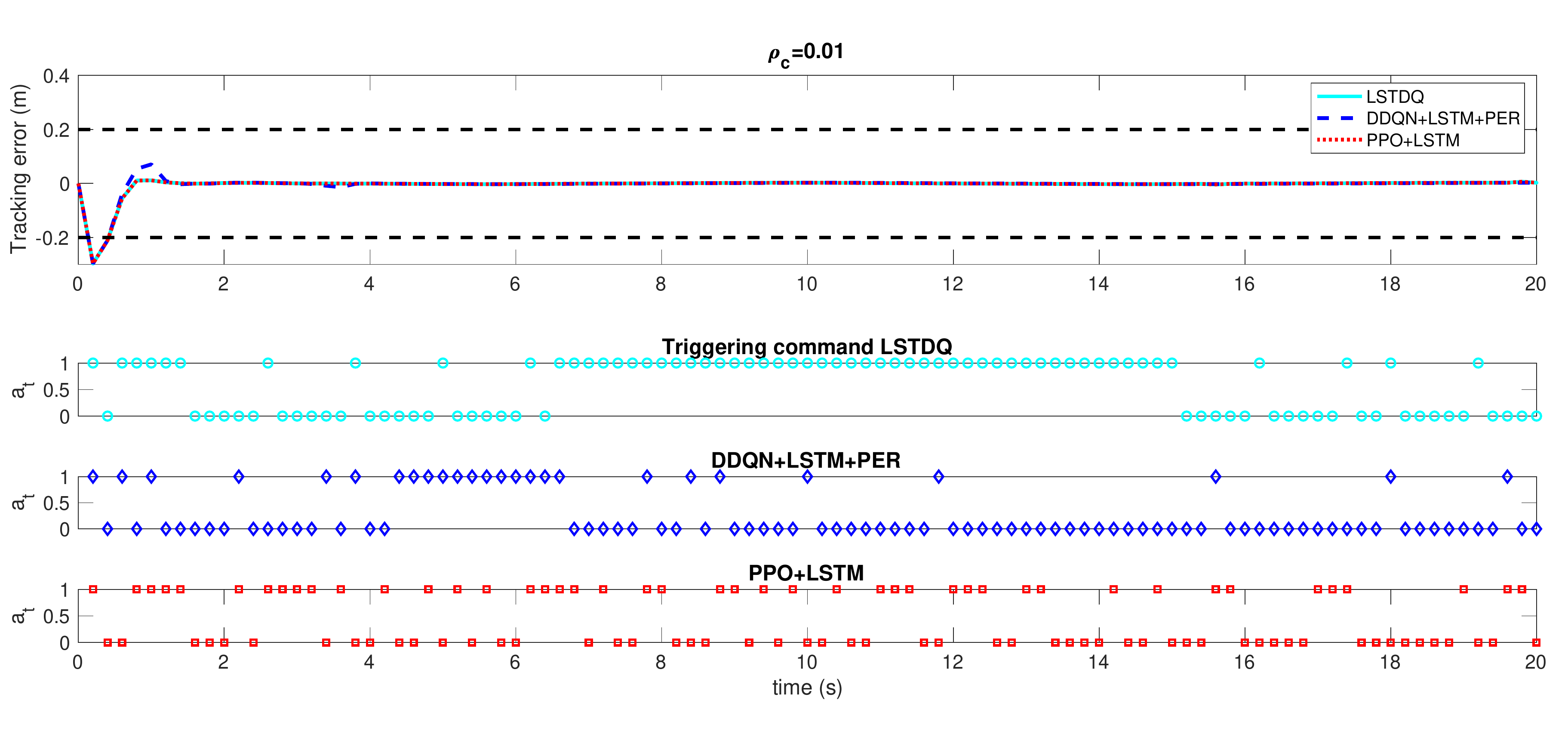}
	\caption{Experimental results of deep-RL-eMPC for the reward function with $\rho_c=0.01$. The comparison of the tracking error using three different RL algorithm in deep-RL-eMPC (first row). The corresponding triggering commands $a_t$ during the process when using LSTDQ (second row), DDQN+LSTM+PER (third row) and PPO+LSTM (fourth row).} 
	\label{fig: DDQN}
\end{figure*}
We train the off-policy RL algorithms over 50,000 steps, which is around 500 episodes, each with a length of $T=20s$ and a sampling time of $dt=200~ms$, i.e., episode horizon is $T_e=100$ time steps. On-policy algorithms, e.g., PPO, often require longer training time but with improved stability \cite{christodoulou2019soft}, thus we train them for 1000 episodes for better convergence. For MDP, we set the discount factor $\gamma = 0.99$ and batch size $N=64$. The learning rate and replay buffer size are set as $\eta=1e-4$ and 5,000, respectively. Also, $\epsilon$-greedy is adopted in DDQN with $\epsilon$ linearly decaying from 1.0 to 0.01 during the first 5000 steps of training.
\subsection{Simulation Results and Analysis}\label{sec:exp}
Numerical simulation results on the evaluation returns for $\rho_c=0$, $0.001$, $0.01$ with the threshold-based benchmark and different variants of RL algorithms are summarized in Tab.~\ref{table:Evaluation return}. The simple linear Q-learning method (least-square temporal difference Q-learning, LSTDQ)~\cite{chen2022reinforcement} is also shown here as a benchmark. To measure the computation burden required by different RL algorithms and MPC, we run the simulation 10000 times and use the average time as the time cost. The results show that the average time cost of MPC is about 0.1s while the average time cost of RL algorithms considered in this paper is about $10^{-6}~s$. In other words, each MPC computation requires $10^{5}$ times more computation than evaluating RL policies, and hence the time cost spent on the decision making of RL algorithms is negligible. So overall speaking, fewer MPC queries will provide less computation burden.

The threshold-based event-trigger policy~\cite{chen2021comparison} depends on a manually-tuned threshold to determine when the event is triggered. However, this method is very sensitive to the tracking error and is susceptible to over-triggering problems when the error is large. This causes the return of the threshold-based method around $1.6$ for all three different $\rho_c$, much worse than the RL-based methods as shown in Tab.~\ref{table:Evaluation return}.

Comparing LSTDQ, SAC, DDQN, and PPO, experimental results clearly show that deep-RL-eMPC frameworks achieve better evaluation return than the the conventional threshold-based approach and previous LSTDQ for all three different $\rho_c$. It is also shown that PPO presents the best result under $\rho_c=0$ and $\rho_c=0.001$, while DDQN performs better when $\rho_c=0.01$ in terms of evaluation return, partly due to the low overestimation. To show the flexibility of the proposed framework, PER buffer and LSTM are employed to foster the exploration and efficiency of the training of DDQN and PPO. PPO is an on policy RL method and PER cannot be applied to this method, so only PPO+LSTM is tested. Specifically, DDQN+LSTM+PER and PPO+LSTM are implemented and compared. The experimental results show that LSTM and PER significantly increase the evaluation return of the system, outperforming the baseline methods. SAC performs well when $\rho_c = 0$, while it fails in the more challenging cases when $\rho_c = 0.001$ or $0.01$. The intrinsic reason for the poor performance of SAC deserves to be investigated in the future work.

Recall that the hyperparameter $\rho_c$ can be used to balance control performance and triggering frequency. When $\rho_c=0$, RL triggers MPC at nearly every time step and achieves the smallest tracking error. As the value of $\rho_c$ increases, the rewards function (\ref{r-nmpc}) penalizes more on triggering MPC, resulting in less frequent events and higher MPC costs $J_\text{mpc}$. The bigger the $\rho_c$ is, the larger penalty the system will give for triggering the events. From Tab.~\ref{table:Evaluation return}, we can see when $\rho_c$ is larger, the system tends to give smaller returns because of the larger punishment of triggering the events. 

Fig.~\ref{fig: LQ} shows the path following error and event triggering command $a_t$ when using three different RL algorithm (LSTDQ, DDQN+LSTM+PER, PPO+LSTM) in the deep-RL-MPC framework when using different $\rho_c$ in reward equation~\ref{Eq: rhoc}. The first row shows the comparison of the tracking error using three different RL algorithm in deep-RL-eMPC. The corresponding triggering commands $a_t$ during the process is showed in second row when using LSTDQ, is showed in third row when using DDQN+LSTM+PER and is showed in fourth row when using PPO+LSTM. The best results from deep-RL-eMPC when $\rho_c=0$ and $\rho_c=0.001$ are from PPO+LSTM and when $\rho_c=0.01$ is from DDQN+LSTM+PER. In LSTDQ, when $\rho_c=0$, $E_{mpc}=0.062$ and the triggering frequency is $0.902$. When $\rho_c=0.001$, $E_{mpc}=0.157$ and the triggering frequency is $0.931$.  When $\rho_c=0.01$, $E_{mpc}=0.66$ and the triggering frequency is $0.559$. In PPO+LSTM, when $\rho_c=0$, $E_{mpc}=0.055$ and the triggering frequency is $0.99$. In this situation, there is no penalty on triggering MPC, and the RL agent triggers MPC for nearly every sampling time, and the path tracking error is the smallest. It results in a triggering frequency of $5~Hz$ as the sampling time is $dt=0.2~s$. When $\rho_c=0.001$, $E_{mpc}=0.059$ and the triggering frequency is $0.594$. In this situation, the RL agent tends to trigger an event when the tracking error is large, and keeps silent when the error is going to be around $0$. When $\rho_c=0.01$, DDQN+LSTM+PER achieves the best performance with $E_{mpc}=0.171$ and the triggering frequency is $0.255$. In this situation, the event-trigger pattern is similar to that of $\rho_c=0.001$, but with a lower triggering frequency. It is worth noting that, for each case, DDQN+LSTM+PER triggers MPC less frequently (resulting in less MPC computation) while incurring smaller MPC cost (resulting in better control performance). We can then conclude that DDQN+LSTM+PER and PPO+LSTM outperforms the previous LSTDQ method as presented in \cite{chen2022reinforcement}.

\section{CONCLUSION}
\label{sec6:CONCLUSION}

This paper investigated a path following problem for autonomous driving. A novel event-triggered model predictive control (eMPC) framework with the triggering policy obtained from deep reinforcement learning was presented to solve the problem. A reward function was proposed to balance control performance and event trigger frequency through a hyper-parameter $\rho_c$. Compared to existing eMPC, the proposed algorithm does not require any knowledge of the closed-loop dynamics (i.e., model-free) and offers better performance. We also show that incorporating techniques such as priority experience replay and long-short term memory can significantly enhance the performance. The learnt deep RL-based triggering policy effectively decreases the computational burden while achieving satisfactory control performance. In future work, we will consider time-varying computational budget and cost using this deep-RL-eMPC framework for autonomous driving path following, as well as other applications of relevance and impact. Additionally, we will also examine the stability and convergence of the proposed deep-RL-eMPC framework.  

\balance
\bibliographystyle{IEEEtran}
\bibliography{references}

\end{document}